\documentclass{article}

\usepackage{arxiv}

\usepackage[utf8]{inputenc} 
\usepackage[T1]{fontenc}    
\usepackage{hyperref}       
\usepackage{url}            
\usepackage{booktabs}       
\usepackage{amsfonts}       
\usepackage{nicefrac}       
\usepackage{microtype}      
\usepackage{lipsum}
\usepackage{graphicx}
\usepackage{amsmath}
\usepackage{amssymb}
\usepackage{pbox}

\title{DeepNovoV2: Better de novo peptide sequencing by deep learning}

\author{
  Rui Qiao\\
  Department of Statistics and Actuarial Science\\
  University of Waterloo\\
  \texttt{rqiao@uwaterloo.ca} \\
  \And
  Ngoc Hieu Tran \\
  David R. Cheriton School of Computer Science\\
  University of Waterloo\\
  \texttt{nh2tran@uwaterloo.ca} \\
  \And
  Lei Xin \\
  Bioinformatic Solution Incorporation \\
  \texttt{lxin@bioinfor.com} \\
  \And
  Baozhen Shan \\
  Bioinformatic Solution Incorporation \\
  \texttt{bshan@bioinfor.com} \\
  \And
  Ming Li \\
  David R. Cheriton School of Computer Science\\
  University of Waterloo\\
  \texttt{nh2tran@uwaterloo.ca} \\
  \And
  Ali Ghodsi\\
  Department of Statistics and Actuarial Science\\
  University of Waterloo\\
  \texttt{ali.ghodsi@uwaterloo.ca} \\
}

\begin{document}
\maketitle

\begin{abstract}
Personalized cancer vaccines are envisioned as the next generation rational cancer immunotherapy. The key step in developing personalized therapeutic cancer vaccines is to identify tumor-specific neoantigens that are on the surface of tumor cells. A promising method for this is through \textit{de novo} peptide sequencing from mass spectrometry data. In this paper we introduce DeepNovoV2, the state-of-the-art model for \textit{de novo} peptide sequencing\footnote{Our pytorch implementation of DeepNovoV2 and the test dataset could be downloaded here: https://github.com/volpato30/DeepNovoV2}. In DeepNovoV2, a spectrum is directly represented as a set of (\textit{m/z}, \textit{intensity}) pairs, therefore it does not suffer from the accuracy-speed/memory trade off problem. The model combines an order invariant network structure (T-Net) and recurrent neural networks and provides a complete end-to-end training and prediction framework to sequence patterns of peptides. Our experiments on a wide variety of data from different species show that DeepNovoV2 outperforms previous state-of-the-art methods, achieving 13.01-23.95 \% higher accuracy at the peptide level.

\end{abstract}

\section{Introduction}

\subsection{Personalized Cancer Vaccines}
The human leukocyte antigen (HLA) system regulates the immune system by bringing short peptides to the surface of cells. These peptides, often referred to as HLA peptides, are produced from digested proteins that are broken down in the proteasomes. When a cell is infected by a virus or grows malignant, non-self HLA peptides (from the virus or mutated proteins) will be presented on the surface of the cell so that T cells could recognize and subsequently kill the cell\cite{bassani2016unsupervised}. For cancer cells, these mutated tumor-specific HLA peptides are referred to as neoantigens.

Based our current understanding of immunology, the generation of an antitumor immune response relies on the activation of professional antigen-presenting cells (APCs)\cite{hu2018towards}. Personalized cancer vaccines are designed to activate APCs, and to hopefully promote an efficient antitumor immune response in the patient's immune system. Neoantigens are peptides that are found on cancer cells and not on normal cells, so they represent ideal targets for cancer vaccines. Up until now, several independent studies have shown successful clinical trials for patients with melanoma\cite{ott2017immunogenic}\cite{sahin2017personalized}\cite{carreno2015dendritic}.
However, a major challenge for personalized cancer vaccine therapy remains the efficient identification and validation of neoantigens (tumor-specific antigens) for each patient\cite{vitiello2017neoantigen}. 
Currently, candidate neoantigens could be identified with a proteogenomics method that involves three steps: First, perform whole exome sequencing and RNA sequencing on the patient's tumor samples to find somatic mutations, and then build a tailored protein database from those identified mutations. Second, perform a tailored database search on the mass spectrums of tumor samples to find candidate tumor specific antigens\cite{bassani2016direct}. Third, the reported candidate neoantigens are further filtered with binding affinity prediction tools like NetMHCpan\cite{jurtz2017netmhcpan}. 

However, Tran et al. (2019) noted that this proteogenomics method might be suboptimal because existing database search engines (used in step two) are not designed for HLA peptides and have bias towards trypic peptides, and because database search engines might have sensitivity and specificity issues when dealing with the large search space created by exome sequencing\cite{tran2019identifying}.
Alternatively, Tran et al. (2019) proposed a workflow that could identify neoantigens directly and solely from mass spectrometry (MS) data. In this workflow, a personalized \textit{de novo} peptide sequencing model is trained by the identified peptide-spectrum matches from a database search result. The trained model is then applied on the unidentified mass spectrums to predict new peptide sequences. The authors reported that their workflow was capable of finding highly confident neoantigens which had not been reported by the proteogenomics method. The success of this novel workflow relies heavily on an accurate and efficient \textit{de novo} peptide sequencing algorithm.

\subsection{Mass Spectrometry (MS)}

MS is a popular and powerful tool for chemical analysis which has contributed in different areas of research including, but not limited to, chemistry, physics and biochemistry\cite{urban2016quantitative}. In MS, samples (typically in liquid or gas form) are loaded into a mass spectrometer which consists of an ion source, a mass analyzer, and an ion detector. The ion source would produce gas phase ions from the sample being studied, the mass analyzer would separate those ions according to their mass-to-charge ratio ($m/z$), and the ion detector would detect the ions and record their relative abundance. Therefore the outputs of a mass spectrometer are spectrums. Each spectrum is a set of $(m/z, intensity)$ pairs where the $intensity$ value represents the abundance of ions with the mass-to-charge ratio $m/z$.

\begin{figure}[h]
    \centering
    \includegraphics[scale=0.65]{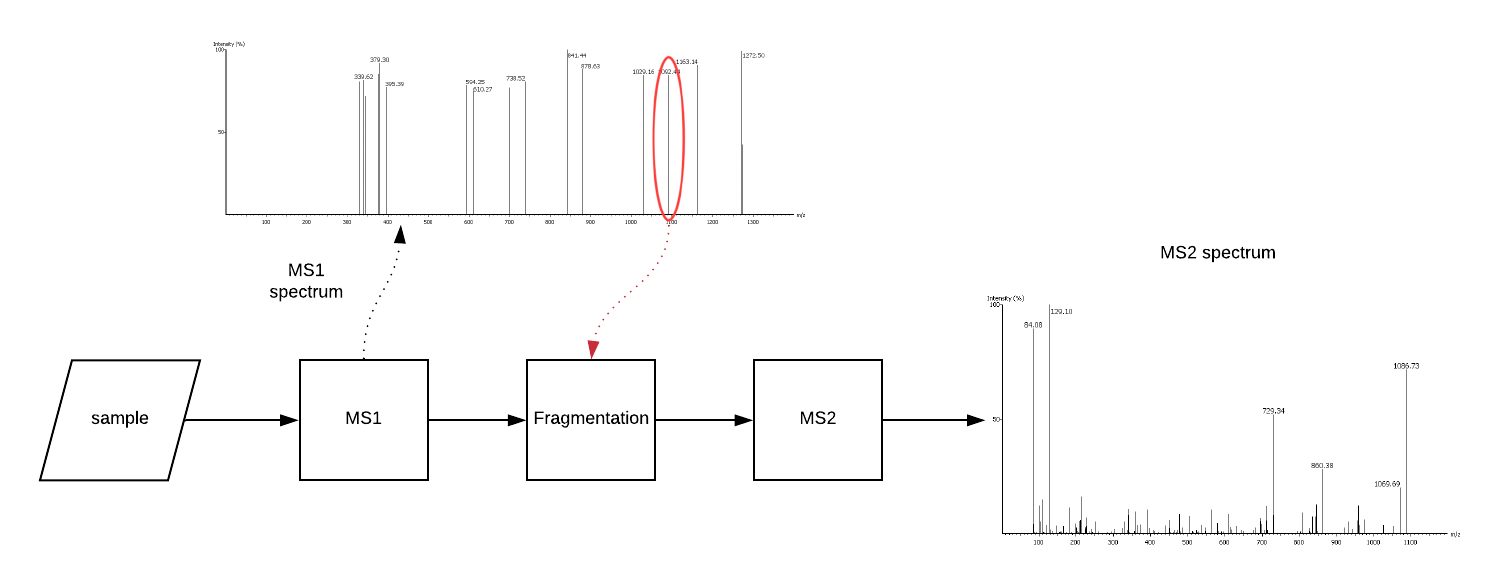}
    \caption{Diagram of tandem mass spectrometry}
    \label{MS/MS}
\end{figure}

Tandem mass spectrometry (MS/MS) is a technique for utilizing two or more different types of mass analyzers to enhance analysis through fragmentation of the input molecules\cite{newton2004plant}. It is widely used in proteomics to analyze protein and peptide sequences. In this approach, biological samples (e.g. tumor tissues, plasma and urine) would be pre-processed with a certain protease (e.g. trypsin and Endoproteinase Lys-C) such that the proteins are cleaved into short peptides. These peptides are then fed into a spectrometer and the output is denoted as MS1 spectrum. Each signal in the MS1 spectrum is called a precursor ion, which typically represents a certain kind of peptide. Next, some precursor ions are selected to be fragmented into smaller pieces called fragment ions. These fragments are further fed into another mass spectrometer. The final output is denoted as MS2 spectrums. Figure \ref{MS/MS} shows the process of MS/MS.

\begin{figure}[h]
    \centering
    \includegraphics[scale=0.37]{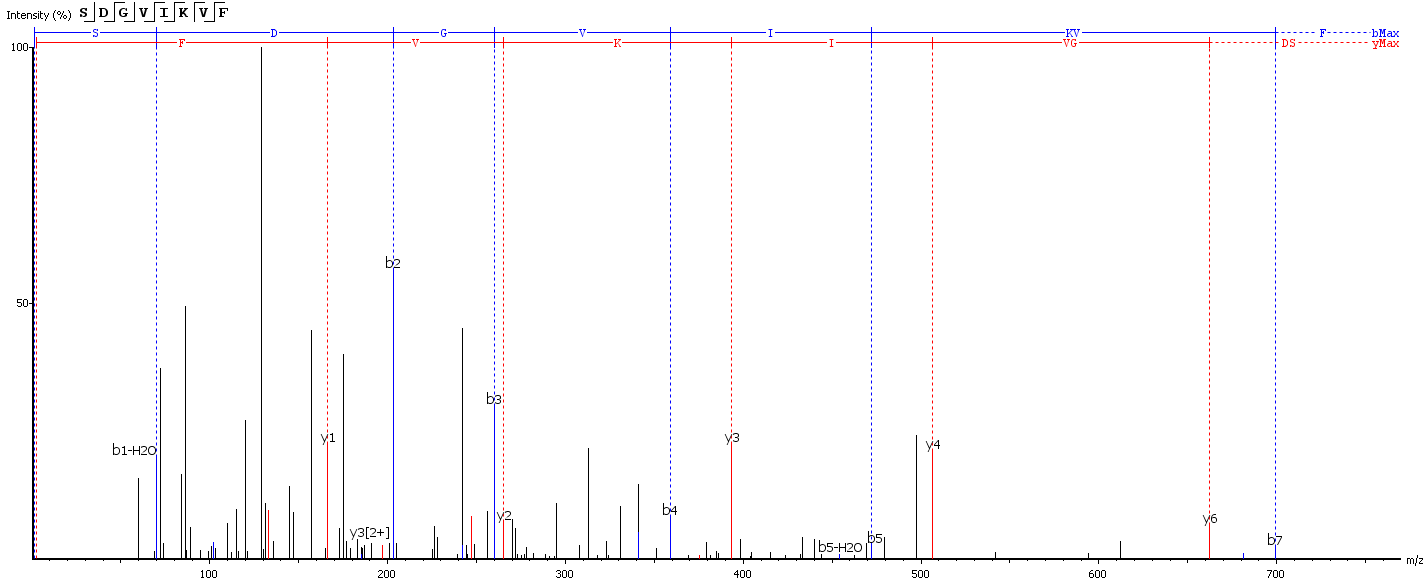}
    \caption{Sample MS2 spectrum}
    \label{sample}
\end{figure}

From the precursor mass (i.e. mass of the peptide) and the fragment ion signals contained in the MS2 spectrum it is possible to recover the exact amino acid sequence of the original peptide. This solution is called \textit{de novo} peptide sequencing. To some extent this challenge is similar to the image captioning problem in computer vision. We need to find a sequence of amino acids (or words), under some constraints (the total mass of the sequence should equal the precursor mass), that best describe the spectrum (or image). A sample MS2 spectrum for peptide "SDGVIKVF" is shown in Figure \ref{sample}.

\subsection{\textit{De novo} Peptide Sequencing}
In proteomics, \textit{de novo} peptide sequencing is the key technology for finding new peptide or protein sequences. It has successful applications in assembling monocolonal antibody sequences (mAbs)\cite{tran2016complete} and great potential in identifying neoantigens for personalized cancer vaccines\cite{tran2019identifying}\cite{tran2019deep}. Given the importance of the \textit{de novo} peptide sequencing technology, substantial research has been done in this area and different tools have been proposed\cite{denovo1999}\cite{denovo2001}\cite{denovo2003}\cite{denovo2005}\cite{denovo20052}. In 2017, Tran et al. first introduced deep learning to \textit{de novo} peptide sequencing and proposed DeepNovo, a neural network based \textit{de novo} peptide sequencing model\cite{tran2017novo} for Data Dependent Acquisition (DDA) MS data. Inspired by the success of image captioning models\cite{imagecaption2015}, DeepNovo integrated two fundamental types of neural networks, Convolutional Neural Networks (CNNs) and Long short-term memory (LSTM), to extract features from both the spectrum and the "language model" of peptides. In DeepNovo, each spectrum is represented as a long intensity vector and CNNs are applied on segments of this vector to extract features and make predictions of the next amino acid. CNNs have been proven as effective tools for pattern recognition in different applications such as image classification, object detection, and sentiment analysis\cite{simonyan2014very}\cite{redmon2016you}\cite{kim2014convolutional}. By applying CNNs to the intensity vector, DeepNovo could learn from the noise spectrum. It is reported that DeepNovo outperformed the decade-long state-of-the-art records of de novo sequencing algorithms (PEAKS\cite{ma2003peaks}) by a large margin of 38.1–64.0\% at the peptide level\cite{tran2017novo}.

In 2019, Tran et al. further extended DeepNovo on Data Independent Acquisition (DIA) MS data and proposed DeepNovo-DIA\cite{tran2019deep}, the first \textit{de novo} peptide sequencing algorithm for DIA MS/MS spectrums. Compared to DDA, DIA data are in general harder to interpret because the spectrum generated by DIA often contains fragment ions from multiple peptides. However, the multiplexity and noise in the DIA spectrums make deep neural networks a more reasonable choice. In DeepNovo-DIA each detected feature is represented by 5 spectrums, where each spectrum is discretized into an intensity vector as in DeepNovo. Thus the input to DeepNovo-DIA is a matrix of shape 5 by the length of intensity vector. DeepNovo-DIA then applies 2D convolution on the input to utilize the information provided by the extra dimension of retention time and hopefully to learn the coeluting patterns. It is also worth noting that Tran et al. reported that by changing cross entropy loss to focal loss\cite{lin2017focal}, they observed a significant improvement on peptide accuracy\cite{tran2019deep}.

In this paper we propose DeepNovoV2: a model combining an order invariant network structure (T-Net) and recurrent neural networks. In addition, this method  utilizes  local dynamic programming to solve the complex optimization task of de novo sequencing.  Our experiments on datasets from a wide variety of species show that DeepNovoV2 outperforms the current state-of-the-art model DeepNovo by a significant margin of at least 13\%, on a peptide level.

In Section 2 of this paper we will explain how we extract features from spectrums and the structure of our proposed model, and in Section 3 we will demonstrate our experiment results on different datasets. 

\section{Methods}
\subsection{Spectrum Representation}
Previously in DeepNovo, spectrums are represented as intensity vectors where each index of the vector represents a small \textit{m/z} bin and the value represents the sum of intensities of all peaks which fall into that bin. This representation of spectrum naturally has the problem of accuracy versus speed/memory trade-off. For example, by default,  DeepNovo use a spectrum resolution of 10 which means every peak within a 0.1 Da \textit{m/z} bin will be merged together and represented as an element of the intensity vector. However, we lose a lot of useful information about the exact location of each peak during the merging. Conversely, if we need to build a more accurate model and increase the resolution, the result will be a significantly longer intensity vector and the model will need more memory and time to be trained. In DeepNovoV2, to solve the accuracy-speed/memory trade off problem, we propose to directly represent the spectrum as a set of \textit{(m/z, intensity)} pairs. For each spectrum we select the top $n$ most intense peaks (by default we choose $n=500$), and represent the spectrum as $\{({m/z}_k, i_k)\}_{k=1}^n$. Further, we denote $\mathbf{M}_{observed}=({m/z}_1, \cdots, {m/z}_n)$ as the observed m/z vector and $\mathbf{i}=(i_1, \cdots, i_n)$.

\subsection{Feature Extraction}
We denote the number of tokens as $n_{vocab}$ and number of ion types as $n_{ion}$. To make a fair comparison with DeepNovo, we keep $n_{vocab}=26$ (20 amino acid residues, three variable modifications, and three special tokens: "start", "end" and "pad") and use the same eight ion types (b, y, b(2+), y(2+), b-H2O, y-H2O, b-NH3, and y-NH3) as in the original implementation of DeepNovo\cite{tran2017novo}. Similar to DeepNovo, at each step we compute the theoretical \textit{m/z} location for each token and ion type pair. The result is a matrix of shape $(n_{vocab}, n_{ion})$ and denoted as $\mathbf{M}_{theoretical}$. Next we expand the dimension of $\mathbf{M}_{observed}$ to make it a 3-dimensional tensor of shape $(n, 1, 1)$, and then repeat $\mathbf{M}_{observed}$ on second dimension for $n_{vocab}$ times and on third dimension for $n_{ion}$ times. The result is denoted as $\mathbf{M}'_{observed}$ and it is a tensor of shape $(n, n_{vocab}, n_{ion})$. Similarly, we expand $\mathbf{M}_{theoretical}$ to the shape of $(1, n_{vocab}, n_{ion})$, repeat on first dimension for $n$ times, and denote the result as $\mathbf{M}'_{theoretical}$. We can then compute the $m/z$ difference tensor (denoted as $\mathbf{D}$) in which each element represents the difference between the $m/z$ value for an observed peak and the theoretical $m/z$ for a token and ion type pair.

\begin{equation}
    \label{d_equation}
    \mathbf{D} = \mathbf{M}'_{observed} - \mathbf{M}'_{theoretical}
\end{equation}

It is worth noting that Equation \ref{d_equation} could be computed efficiently by using the "broadcast" behaviour in popular frameworks like Tensorflow\cite{tensorflow} and PyTorch\cite{pytorch}. 

\begin{equation}
    \label{sigma_equation}
    \sigma(\mathbf{D}) = \exp\{-|\mathbf{D}| * c\}
\end{equation}

Based on expert knowledge of \textit{de novo} peptide sequencing, we designed an activation function $\sigma$ shown in Equation \ref{sigma_equation}. Here the exponential and absolute operations are all element-wise operations. The intuition for $\sigma$ is that an observed peak could only be considered matching a theoretical \textit{m/z} location if the absolute \textit{m/z} difference is small. For example, if we set $c=100$, then an observed peak that is 0.02 Da away from a theoretical location would generate a signal of $e^{-2} \approx 0.135$, which is only one seventh of the signal of a perfect match. In our experiments, we tried setting $c$ to be a trainable parameter and updating it through backpropagation. It shows similar performance with setting $c=100$;  therefore,  for the simplicity of the model, we choose to fix $c$ to be $100$. 
\begin{equation}
    \label{f_equation}
    \mathbf{F} = \sigma(\mathbf{D})' \oplus \mathbf{I} 
\end{equation}
Finally, the feature matrix $\mathbf{F}$ for prediction is simply the concatenation of $\sigma(D)$ and $\mathbf{i}$, as shown in Equation \ref{f_equation}. Here $\sigma(D)'$ is a $n$ by $n_{vocab} \times n_{ion}$ matrix reshaped from $\sigma(D)$, $\mathbf{I}$ is a $N$ by $1$ matrix reshaped from $\mathbf{i}$ and $\oplus$ represents concatenation along the second dimension.

\subsection{T-Net}
\begin{figure}[h]
    \centering
    \includegraphics[scale=0.75]{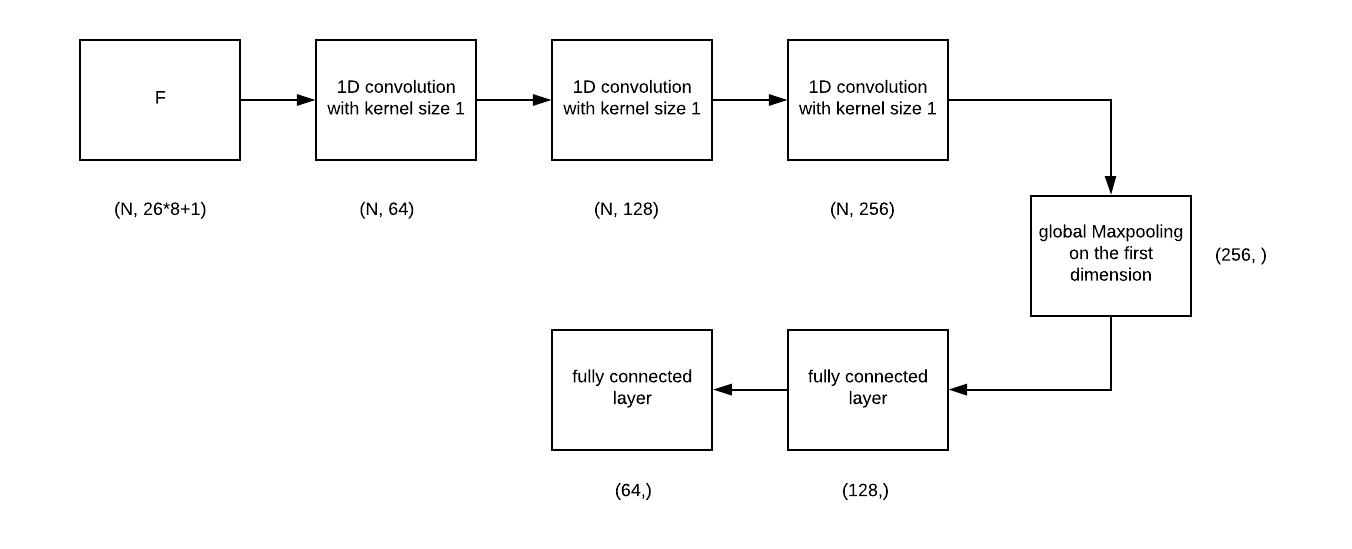}
    \caption{T-Net}
    \label{tnet}
\end{figure}

\begin{figure}[h]
    \centering
    \includegraphics[scale=0.7]{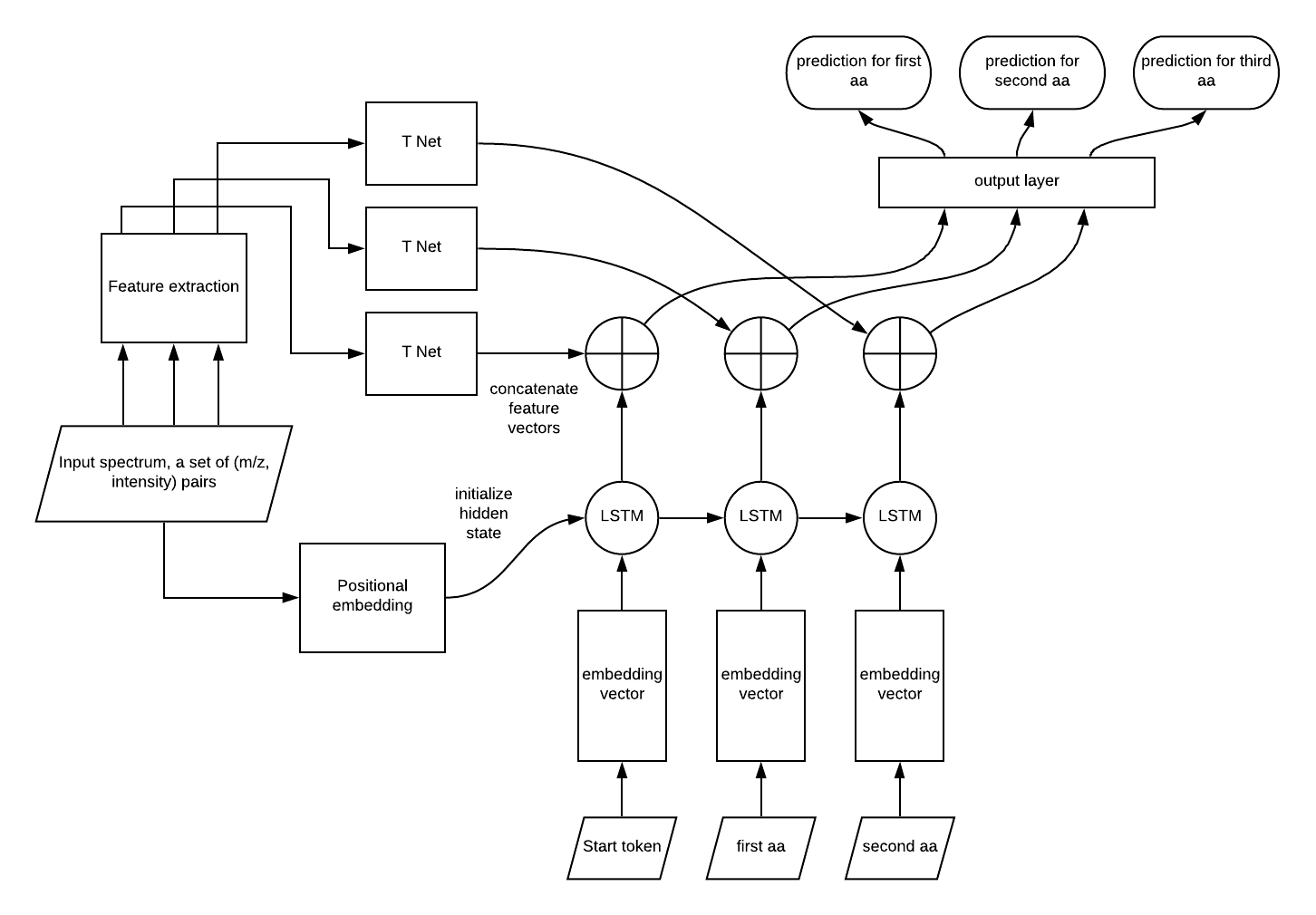}
    \caption{Structure of DeepNovoV2}
    \label{dnv2}
\end{figure}

A spectrum is \textbf{set} of $(m/z, intensity)$ pairs, which means the order of peaks should be irrelevant. Therefore, the prediction network should have an order invariant property with respect to the first dimension of $F$. To the best of our knowledge, T-Net (the building block of Point Net\cite{qi2017pointnet}) is the first model designed for this kind of order invariant data, so we choose to use it to process $F$. In essence, T-Net is composed of three 1d convolutional layers with kernel size one, followed by a global max pooling layer and three fully connected layers. As with DeepNovo\cite{tran2017novo}, we could add a LSTM aside from the T-net to make use of the language model information of peptides. The structure for T-Net is shown in Figure \ref{tnet} and the full model structure of DeepNovoV2 is shown in Figure \ref{dnv2}. As suggested by Tran et al.\cite{tran2019deep}, we used focal loss\cite{lin2017focal} instead cross entropy loss when training the model.

\subsection{Initial State for LSTM}

Similar to DeepNovo\cite{tran2017novo}, we could include a LSTM module to capture the "language model" for peptides. To make good predictions about the next amino acid, it is important for the LSTM to be initialized with information about the original spectrum. DeepNovo used a spectrum CNN to extract features from intensity vectors and then used the extracted features to initialize the LSTM. In DeepNovoV2, we replaced the spectrum CNN structure with a simple embedding matrix. Specifically, we created a sinusoidal m/z positional embedding, as suggested by Vaswani et al\cite{vaswani2017attention}:

\begin{align*}
    PE_{(loc, 2k)} & = sin(loc/10000^{\frac{2k}{d_{lstm}}}) \\
    PE_{(loc, 2k+1)} & = cos(loc/10000^{\frac{2k}{d_{lstm}}})
\end{align*}

Here $loc$ represents the m/z location after discretization, and $d_{lstm}$ represents the dimension of the LSTM module. The sinusoidal embedding has a desired property that any distance $m$, $PE_{loc+m}$ could be represented as a linear function of $PE_{loc}$\cite{vaswani2017attention}. This property is important because in mass spectrums the m/z differences between observed peaks contains useful information that indicates which amino acids could possibly exist. By using the sinusoidal positional embeddings, we anticipate that the model could easily learn to extract information from the m/z difference between peaks.

For each spectrum $\{({m/z}_k, i_k)\}_{i=1}^n$, we denote the positional embedding for ${m/z}_k$ as $\mathbf{e}_k$. The spectrum representation $\mathbf{s}$ is computed by:

\begin{equation}
\mathbf{s} = \sum_{k=1}^n i_k \mathbf{e}_k
\end{equation}

Next, the initial hidden states of the LSTM module (i.e. $h_0$ and $c_0$) will be initialized to $S$. By replacing spectrum CNN to a fixed positional embedding matrix, we reduced the number of parameters and computation needed in the model. Our experiment results show that initializing with $\mathbf{s}$ gives similar results when comparing to initializing with the result of spectrum CNN.

\subsection{Training Parameters}
We train DeepNovoV2 with Adam algorithm\cite{kingma2014adam} with an initial learning rate of $10^{-3}$. After every 300 training steps, the loss on validation set is computed. If the validation loss has not achieved a new low in the most recent ten evaluations, then the learning rate is dropped by half.

\section{Results and Analysis}

\subsection{Evaluation Metrics}

To evaluate the performance of \textit{de novo} sequencing models, we use the same metrics as proposed by DeepNovo\cite{tran2017novo}. A predicted amino acid is considered "matched" with a real amino acid if their mass difference is less than 0.1 Da and if the prefix masses before them are different by less than 0.5 Da. The amino acid level recall (or precision) is defined as the ratio of the total number of matched amino acids over the total number of amino acids in real peptide sequences (or predicted peptide sequences). And the peptide level recall denotes the fraction of real peptide sequences that are fully correctly predicted. 

\subsection{Experiment Results}

\begin{table}
\centering
\begin{tabular}{|l|l|l|l|l|}
\hline
                    & \pbox{10cm}{DeepNovo \\ without lstm} & \pbox{10cm}{DeepNovo \\ with lstm} & \pbox{10cm}{DeepNovoV2 \\ without lstm} & \pbox{10cm}{DeepNovoV2 \\ with lstm} \\ \hline
amino acid recall & $66.44\% (\pm 0.18\%)$                & $67.81\% (\pm 0.21\%)$             & $71.59\% (\pm 0.18\%)$                  & $72.46\% (\pm 0.15\%)$              \\ \hline
amino acid precision & $66.42\% (\pm 0.22\%)$                & $67.67\% (\pm 0.25\%)$             & $71.65\% (\pm 0.24\%)$                  & $72.25\% (\pm 0.20\%)$               \\ \hline
peptide recall    & $30.33\% (\pm 0.15\%)$               & $32.96\% (\pm 0.31\%)$             & $38.10\% (\pm 0.06\%)$                  & $39.24\% (\pm 0.29\%)$               \\ \hline
\end{tabular}
\caption{Amino acid recall, amino acid precision and peptide recall on ABRF DDA data}
\label{t1}
\end{table}

We tested our model on a DDA dataset of Hela samples\footnote{PRG 2018: Evaluation of Data-Independent Acquisition (DIA) for Protein Quantification in Academic and Core Facility Settings. https://abrf.org/research-group/proteomics-research-group-prg} (denoted as ABRF dataset). The train-valid-test split is done in the same way suggested by DeepNovo\cite{tran2017novo}, i.e. train, valid and test sets do not share any common peptides. Both DeepNovo and DeepNovoV2 models are trained for 20 epochs. The weights with the smallest validation error are then selected to do beam search. We also use the knapsack dynamic programming algorithms as described by Tran et al. (2017) to limit the search space\cite{tran2017novo}. 

The mean value and standard deviation of metrics from 5 independent runs on ABRF dataset are shown in Table \ref{t1}. The results show that DeepNovoV2 outperforms DeepNovo by  $6.85\%$  on amino acid recall, $6.76\%$ on amino acid precision and $19.05\%$ on peptide recall.

We also tested DeepNovoV2 with the cross species data published by Tran et al.\cite{tran2017novo} (MSV000081382). This data set contains DDA mass spectrums from nine species. For each species we train a model using spectrums from the other 8 species and then test the model. In these experiments we set $n=1000$.

\begin{figure}[h]
    \centering
    \includegraphics[scale=0.27]{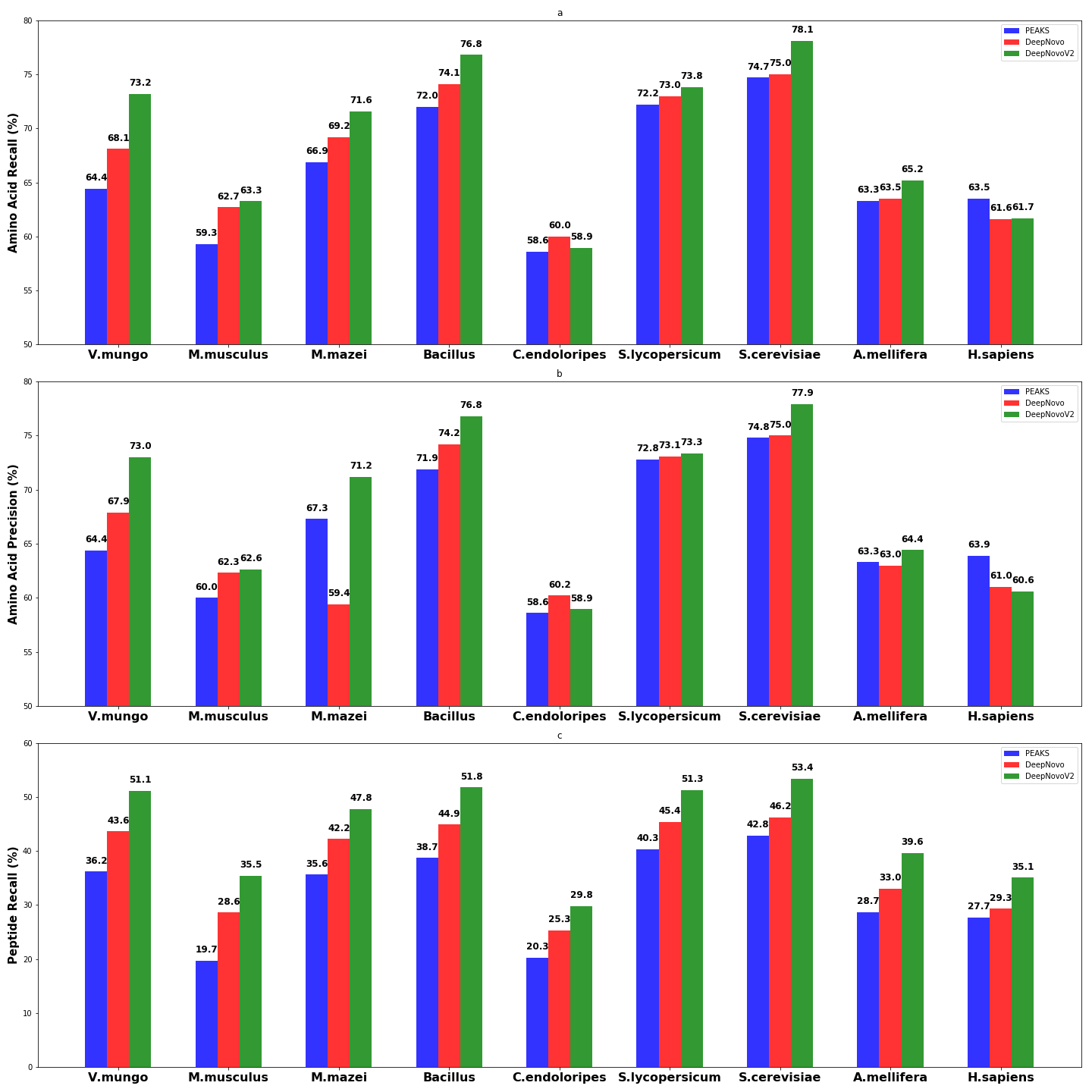}
    \caption{Amino acid recall, amino acid precision and peptide recall of DeepNovo and DeepNovoV2}
    \label{comparison}
\end{figure}

As we can see from Figure \ref{comparison}, DeepNovoV2 outperforms DeepNovo consistently on peptide recall, by a large margin of 13.01–23.95\%.

\section{Conclusion and Future Research}
We propose DeepNovoV2: a neural network based \textit{de novo} peptide sequencing model that outperforms the previous state-of-the-art model by a significant margin (at least $13\%$ on peptide recall). This improvement is achieved through the change in spectrum representations and the use of an order invariant network structure. We believe these techniques could be applied to other important problems in MS such as database search or feature detection, and we continue to research in this direction.

\bibliographystyle{unsrt}  
\bibliography{references}  

\end{document}